\documentclass[letterpaper, 10 pt, conference]{ieeeconf}  

\IEEEoverridecommandlockouts                              

\usepackage{graphics} 
\usepackage{epsfig} 
\usepackage{times} 
\usepackage{amsmath} 
\usepackage{amssymb}  
\usepackage{caption}
\usepackage{subcaption}
\usepackage{tikz}
\usepackage{comment}

\newcommand{\dfb}{\stackrel{\Delta}{=}}
\newcommand{\R}{\ensuremath{\mathbb R}}

\title{\LARGE \bf
Multi-robot motion-formation distributed control with sensor self-calibration: experimental validation.
}

\author{Hector Garcia de Marina$^{1}$ Johan Siemonsma$^{2}$ Bayu Jayawardhana$^2$ and Ming Cao$^2$
\thanks{$^{1}$H. Garcia de Marina is with the Unmanned Aerial Systems Center, M{\ae}rsk Mc-Kinney M{\o}ller Institute, Southern University of Denmark, Denmark. {\tt\small hgm@mmmi.sdu.dk}.}%
\thanks{$^{2}$J. Siemonsma, B. Jayawardhana and M. Cao are with the Engineering and Technology Institute of Groningen, Faculty of Science and Engineering, University of Groningen, the Netherlands. {\tt\small \{b.jayawardhana,m.cao\}@rug.nl}.}%
}
\begin{document}

\maketitle
\thispagestyle{empty}
\pagestyle{empty}

\begin{abstract}
In this paper, we present the design and implementation of a robust motion formation distributed control algorithm for a team of mobile robots. The primary task for the team is to form a geometric shape, which can be freely translated and rotated at the same time. This approach makes the robots to behave as a cohesive whole, which can be useful in tasks such as collaborative transportation. The robustness of the algorithm relies on the fact that each robot employs only local measurements from a laser sensor which does not need to be off-line calibrated. Furthermore, robots do not need to exchange any information with each other. Being free of sensor calibration and not requiring a communication channel helps the scaling of the overall system to a large number of robots. In addition, since the robots do not need any off-board localization system, but require only relative positions with respect to their neighbors, it can be aimed to have a full autonomous team that operates in environments where such localization systems are not available. The computational cost of the algorithm is inexpensive and the resources from a standard microcontroller will suffice. This fact makes the usage of our approach appealing as a support for other more demanding algorithms, e.g., processing images from onboard cameras. We validate the performance of the algorithm with a team of four mobile robots equipped with low-cost commercially available laser scanners.
\end{abstract}

\section{INTRODUCTION}
In recent years, the use of individual and independent robots executing repetitive tasks has gradually been replaced by the exploitation of teams of co-operative robots solving multiple tasks together. The usage of robots in a coordinated fashion is already a reality in many tasks such as the transportation of objects \cite{wang2016multi}, area exploration, and environmental surveillance \cite{yuan2010cooperative}. All these tasks have practical applications in today's society such as rescue missions in disaster areas and precision agriculture. In addition, there has been much research in more fundamental topics in the control of multi-agent robotic systems \cite{oh2015survey,olfati2007consensus,AnYuFiHe08,de2016distributed}, which enhances the efficiency and robustness of the successful execution of the previously mentioned tasks. Difficulties arise, however, when an attempt is made to implement such fundamental control methodologies in actual robots. For example, previous published works prefer to demonstrate their algorithms by employing external motion capture systems, not only as a ground truth, in order to localize their teams of robots \cite{mulgaonkar2018robust,wang2017safe,pickem2017robotarium}. While these systems have advantages for rapid prototyping, they present potential problems if one wants to scale up the multi-robot system or to make it independent of the environment.

In formation control algorithms, it is usually assumed that perfect measurement information is available for the robots, and their robustness against noises in the information is guaranteed by the (locally exponential) stability properties of the closed-loop system \cite{sun2016exponential}. However, in multi-robot systems whose task is to maintain a formation shape by using the popular gradient-descent control \cite{KrBrFr08}, if neighboring robots do not have perfectly calibrated sensors, then the exponential robustness of the desired shape does not lead to the stability of the group formation. In particular, the formation will exhibit an undesired collective motion in addition to the distortion of the desired shape \cite{mou2016undirected}. This difficulty can be overcome if the two neighboring robots do not share the same responsibility, e.g., only one robot controls the error distance between two neighbors. This is exemplified in the recent work undertaken by ETH Z{\"u}rich regarding collaborative transportation by rotorcraft \cite{tagliabue2017robust}, where the robots are in a master-slave configuration while non-external localization system is employed. However, leader-follower configuration leads to the \emph{directed topology} that describes the relation between neighboring robots \cite{BaArWe11}, which makes the analysis of the key properties of the formation control algorithm, such as stability, the region of attraction or the convergence time, more complicated once we scale up the number of robots. On the other hand, the mentioned analysis is more tractable when both neighboring robots fully cooperate in an \emph{undirected topology} \cite{BaArWe11}, although one practical drawback of this approach lies in the extra cost of calibrating all pair of sensors on neighboring robots when we increase the number of robots in the team.

In the scenario when two neighboring robots disagree about the distance to be controlled between them, we have proposed in \cite{garcia2015controlling} a solution by adding local estimators to the existing distributed gradient-based formation control law. In this paper we are going to show and experimentally validate that this solution can also be employed for online sensor calibration for a particular definition of a distance error signal.

Further work has led the systematic understanding of how the disagreement in the distances to be controlled by the robots contributes to the collective motion, which eventually has resulted in a new approach for the manipulation, such as coordinated motion, of rigid formations \cite{de2016distributed,de2017taming}. However, to the best knowledge of the authors of this article, there are not works showing the effectiveness of these algorithms in a \emph{fully} distributed and autonomous setup, i.e., where one does not employ any external localization system in the control loop (previously we were simulating local measurements from readings of a global localization system \cite{garcia2015controlling,de2016distributed}), and all the local computations take place on the robots without any communication among them or with a central computer. The experimental validation of such a fully distributed system is crucial in order to assure that no other real issues such as non-synchronized clocks among robots, have a substantial impact on the robustness and performance of the team of robots. This is a relevant subject after the findings showing that the exponential convergence of the gradient-based formation control does not really protect the system against small disturbances in the range sensors. 

The paper is organized as follows. Firstly, we explain in Section \ref{sec: platform} our experimental multi-robot platform equipped with low-cost laser scanners without any calibration, just \emph{out-of-the-box}. Secondly, we introduce in Section \ref{sec: rigid} some notation and the concept of rigid formations for controlling shapes via the popular gradient descent. Thirdly, we show in Section \ref{sec: comp} how an external operator can drive the shape formed by the team of robots as a single entity. In particular, we experimentally show how this formation movement can be precisely achieved with the online calibration, and without requiring any communication between the robots. In addition, we will also present the practical impact of running the formation control law without any calibration routine. Finally, the paper is finished by summarizing some conclusions in Section \ref{sec: con}.

\section{Multi-robot fully distributed system}
\label{sec: platform}
The setup for the experimental verification of the algorithms consists of four mobile robots. The dimensions and mobility of our robots are quite similar to the Kuka Youbot \cite{bischoff2011kuka}. The base of each robot is an aluminum chassis with four 100 mm aluminum Mecanum wheels and a suspension system at the back. This kind of omnidirectional wheels allows the usage of algorithms that focus on kinematic points as motion model as commonly assumed in the literature \cite{oh2015survey}. The maximum speed of the robots is about 1.0 m/s. For the proposed algorithms and their applications, this speed gives us enough room for the control actions without saturating the motors.

Although our robots are equipped with several kinds of sensors, the only source of information employed by the experiments in this paper is a RP-LIDAR laser scanner mounted over their chassis. These laser scanners are employed to measure the relative position of a robot with respect to its neighbors. The experiments aim at achieving formation shapes where the robots are typically separated by around a couple of meters. The employed laser scanners offer an accuracy up to 0.2\% of the measured distance, a maximum range of 6m, and they cover 360 degrees in 0.2 seconds with a resolution of about 1 degree. The motors and the laser scanners are driven by an ATMega microcontroller, which executes the formation control algorithm at a frequency of 5Hz.

There is also onboard of each robot an embedded computer running an Ubuntu 14.04, whose main purpose is to log the experimental data, and to provide a comfortable way to wireless communicate with the robots from a laptop in the same network, e.g., for extracting the logs. In fact, although all the algorithms will be executed in a fully distributed way, an external operator will set the \emph{high level} objectives of the formation, i.e., the operator will command the size and/or the linear and rotational velocities around the centroid of a desired shape. In particular, the operator only needs to send these high level commands, e.g., with a joystick, if he desires to change the current motion of the formation. As it has been mentioned, once the high level command is set, the robots do not exchange or share any kind of communication or information for achieving the motion task.

\begin{figure}
\centering
\begin{subfigure}{.49\columnwidth}
  \centering
  \includegraphics[width=\linewidth]{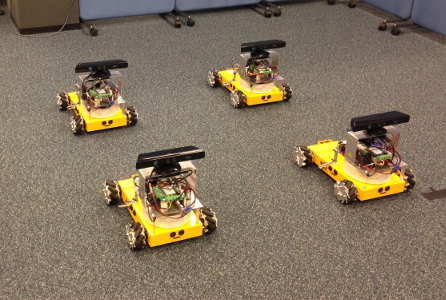}
  \caption{The four robots employed with omnidirectional wheels.}
  \label{fig:su_1}
\end{subfigure}
\begin{subfigure}{.49\columnwidth}
  \centering
  \includegraphics[width=\linewidth]{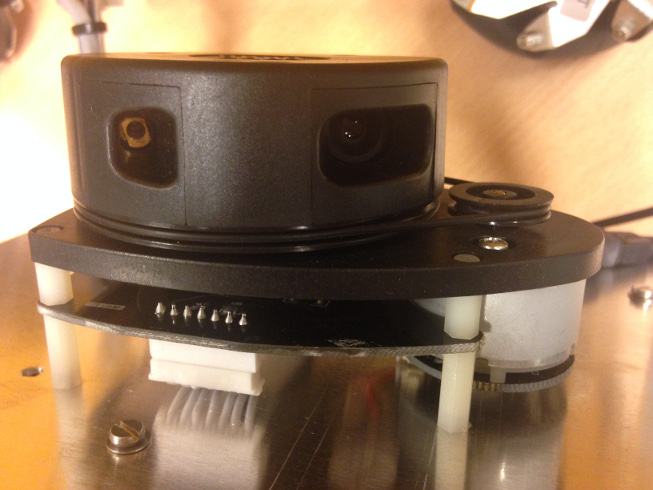}
   \caption{RP-LIDAR laser scanner mounted at the front of each robot.}
  \label{fig:su_4}
\end{subfigure}%
\caption{Hardware used during the experiments.}
\label{fig:labsetup}
\end{figure}

\section{Control of rigid formations}
\label{sec: rigid}
This section introduces and explains the notation and mathematical concepts that will be used throughout the rest of the paper. Consider a team of $n\geq 2$ robots and denote by $p_i\in\mathbb{R}^2, i\in\{1,\dots,n\}$ their 2D positions with respect to some arbitrary and fixed frame of coordinates. The formation control algorithms generate a velocity signal to be tracked by the robots, i.e., the motion of the robots with omnidirectional wheels can be modeled by
\begin{equation}
\dot p = u,
\end{equation}
where $p\in\mathbb{R}^{2n}$ is the stacked vector of positions and $u\in\mathbb{R}^{2n}$ is the generated stacked vector of velocities to be tracked.

A robot does not need to measure its relative position with respect to all the robots in the team, but only with respect to its \emph{neighbors}. The neighbors' relationships are described by an undirected graph $\mathbb{G} = (\mathcal{V}, \mathcal{E})$ with the vertex set $\mathcal{V} = \{1, \dots, n\}$ and the ordered edge set $\mathcal{E}\subseteq\mathcal{V}\times\mathcal{V}$. The set $\mathcal{N}_i$ of the neighbors of robot $i$ is defined by $\mathcal{N}_i\dfb\{j\in\mathcal{V}:(i,j)\in\mathcal{E}\}$. We define the elements of the incidence matrix $B\in\R^{|\mathcal{V}|\times|\mathcal{E}|}$ for  $\mathbb{G}$ by
\begin{equation}
	b_{ik} \dfb \begin{cases}+1 \quad \text{if} \quad i = {\mathcal{E}_k^{\text{tail}}} \\
		-1 \quad \text{if} \quad i = {\mathcal{E}_k^{\text{head}}} \\
		0 \quad \text{otherwise}
	\end{cases},
	\label{eq: B}
\end{equation}
where $\mathcal{E}_k^{\text{tail}}$ and $\mathcal{E}_k^{\text{head}}$ denote the tail and head nodes, respectively, of the edge $\mathcal{E}_k$, i.e., $\mathcal{E}_k = (\mathcal{E}_k^{\text{tail}},\mathcal{E}_k^{\text{head}})$. For undirected graphs, how one sets the direction of the edges is not relevant for the stability results or for the practical implementation of the algorithm \cite{oh2015survey}.

The stacked vector of the sensed relative positions by the robots can be calculated as
\begin{equation}
	z = (B^T \otimes I_2)p,
\end{equation}
where $I_2$ is the $2\times 2$ identity matrix, and the operator $\otimes$ denotes the Kronecker product. Note that each vector $z_k = p_i - p_j$ stacked in $z$ corresponds to the relative position associated with the edge $\mathcal{E}_k = (i, j)$. 

The presented formation control algorithms are based on the distance-based approach, i.e., we are defining shapes by only controlling distances between neighboring robots. These shapes are based on the rigidity graph theory \cite{AnYuFiHe08}. Concretely, the shapes are from a particular class of rigid formations that are \emph{infinitesimally rigid}. A \emph{framework} is defined by the pair $(\mathbb{G}, p)$, where a position is assigned to each node of the graph. Roughly speaking, a framework $(\mathbb{G}, p)$ is infinitesimally rigid in 2D if it is not possible to smoothly move one node of the framework without moving the rest while maintaining constant all the inter-node distances, the framework is invariant under and only under translations and rotations, and the nodes in the framework are not all collinear.

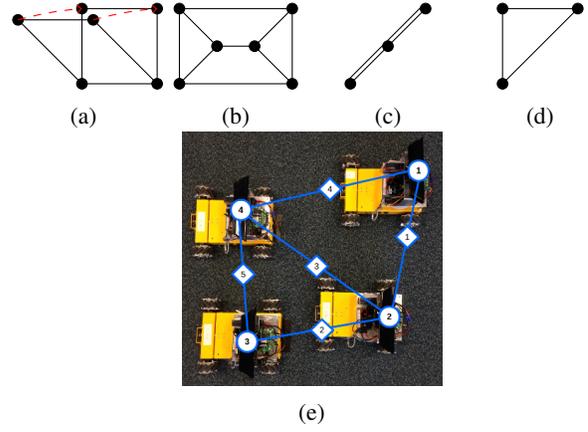
\begin{figure}
	\centering
	\begin{subfigure}{0.22\columnwidth}
		\centering
\begin{tikzpicture}[line join=round]
\draw(0,0)--(1,0)--(1,1)--(0,1)--(0,0);
\filldraw(0,0) circle (2pt);
\filldraw(1,0) circle (2pt);
\filldraw(1,1) circle (2pt);
\filldraw(0,1) circle (2pt);
\filldraw(-.85,.85) circle (2pt);
\filldraw(.15,.85) circle (2pt);
\draw(0,0)--(-.85,.85);
\draw(1,0)--(.15,.85);
\draw(-.85,.85)--(.15,.85);
\draw[red,->,style=dashed] (-.85,.85) to [bend left=5] (0,1);\draw[red,->,style=dashed] (.15,.85) to [bend left=5] (1,1);\end{tikzpicture}
		\caption{}
	\end{subfigure}
	\begin{subfigure}{0.22\columnwidth}
		\centering
\begin{tikzpicture}[line join=round]
\filldraw(0,0) circle (2pt);
\filldraw(0,1) circle (2pt);
\filldraw(.5,.5) circle (2pt);
\filldraw(1,.5) circle (2pt);
\filldraw(1.5,1) circle (2pt);
\filldraw(1.5,0) circle (2pt);
\draw(0,0)--(0,1);
\draw(0,1)--(.5,.5);
\draw(.5,.5)--(0,0);
\draw(1,.5)--(1.5,1);
\draw(1.5,1)--(1.5,0);
\draw(1.5,0)--(1,.5);
\draw(0,1)--(1.5,1);
\draw(.5,.5)--(1,.5);
\draw(0,0)--(1.5,0);
\end{tikzpicture}
		\caption{}
	\end{subfigure}
	\begin{subfigure}{0.22\columnwidth}
		\centering
\begin{tikzpicture}[line join=round]
\filldraw(0,0) circle (2pt);
\filldraw(.5,.5) circle (2pt);
\filldraw(1,1) circle (2pt);
\draw(0,0)--(.5,.5);
\draw(.5,.5)--(1,1);
\draw(-.02,.06)--(.98,1.06);
\end{tikzpicture}
		\caption{}
	\end{subfigure}
	\begin{subfigure}{0.22\columnwidth}
		\centering
\begin{tikzpicture}[line join=round]
\filldraw[fill=white](0,0)--(1,1)--(0,1)--cycle;
\filldraw(0,0) circle (2pt);
\filldraw(1,1) circle (2pt);
\filldraw(0,1) circle (2pt);
\end{tikzpicture}
		\caption{}
	\end{subfigure}
	\begin{subfigure}{0.4\columnwidth}
		\centering
		\includegraphics[width=\linewidth]{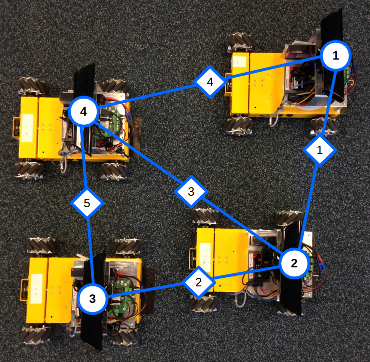}
		\caption{}
	\end{subfigure}
	\caption{a) The square without an inner diagonal is not rigid since we can smoothly move the top two nodes keeping the other two fixed without breaking the distance constraints; b) A rigid but not infinitesimally rigid framework. If we rotate the left inner triangle, then the right inner triangle can be counter-rotated in order to keep constant all the distances; c) A rigid but not infinitesimally rigid framework since the nodes' positions are collinear; d) The triangle is infinitesimally rigid. The only possible transformations in order to keep the distances constant are the translation or rotation of the whole framework; e) An infinitesimally rigid framework. The number in the circles denotes the label of the node or robot $i$ and the number in a diamond represents the edge $k$. Note that robots $1$ and $3$ are not neighbors. This configuration corresponds to the incidence matrix in (\ref{eq: B}).}
	\label{fig: rigid}
\end{figure}

The introduced concepts and notations are illustrated in Figure \ref{fig: rigid}. In particular, throughout the paper the experimental setup consists of four robots with the following incidence matrix defining the neighbors' relationship
\begin{equation}
	B = \begin{bmatrix}
	1 & 0 & 0 & 1 & 0 \\
	-1 & 1 & 1 & 0 & 0 \\
	0 & -1 & 0 & 0 & 1 \\
	0 & 0 & -1 & -1 & -1
	\end{bmatrix}.
\label{eq: B}
\end{equation}

Let $d\dfb\{d_1,\dots,d_k\}, k\in\{1,\dots,|\mathcal{E}|\}$ be a collection of fixed distances, associated to their corresponding edges, which defines locally a desired infinitesimally rigid shape. Then, the error signals to be minimized are given by
\begin{equation}
	e_k(t) := ||z_k(t)|| - d_k,
	\label{eq: e}
\end{equation}
which is slightly different to the traditional $e_k(t) := ||z_k(t)||^2 - d_k^2$ in the literature \cite{oh2015survey}. While the latter makes the closed loop system easier to analyze, it is (\ref{eq: e}) which makes a difference for our proposed online calibration.

The control action for each robot in order to stabilize the desired shape can be derived from the gradient descent of the potential function involving all the error distances to be minimized
\begin{equation}
	V =\frac{1}{2} \sum_{k=1}^{|\mathcal{E}|} (||z_k(t)|| - d_k)^2, \label{eq: Vkquad}
\end{equation}
which leads to the following control action for each robot $i$
\begin{equation}
	^iu_i = -\sum_{j\in\mathcal{N}_i}\frac{^i(p_i-p_j)}{||p_i-p_j||}(||p_i-p_j|| - d_{(i,j)}), \label{eq: ui}
\end{equation}
where each desired distance $d_{(i,j)} = d_{(j,i)}$ is associated with its corresponding $d_k$ for the $k$'th edge of the undirected graph, and the superscript $i$ over the vectorial quantities is used for the representation of a vector with respect to the local frame of coordinates of robot $i$. In fact, one appealing property of the distance-based approach is that robots do not need to share any common orientation \cite{oh2015survey}, i.e., it is irrelevant how the laser scanners are mounted with respect to the others. Indeed, this fact add extra robustness to the proposed formation control law. Note that the laser scanners can measure independently the two terms of each element of the sum in (\ref{eq: ui}), i.e., the relative orientation $\hat z_k =\frac{^i(p_i-p_j)}{||p_i-p_j||}$ and the actual inter-robot distance $||z_k|| =||p_i-p_j||$. Before starting the experiments, the robots roughly know \emph{a priori} where their neighbors are placed. Indeed, this condition can be relaxed if we count on an extra more sophisticated localization system based on vision. In fact, we propose to support such localization systems with the presented algorithm in this paper, since algorithms based on vision can help to identify neighbors but they are computationally expensive, and need of a more complex hardware than a microcontroller. Although we assume that there will not be obstacles between the robots as in Figure \ref{fig: rigid}e), this can also be relaxed by considering switching topologies \cite{oh2015survey}, e.g., some links are missing during a finite time.  As a result of the mentioned assumptions, the robots can obtain and identify straightforwardly their relative positions with respect to their neighbors employing only a laser, and hence they are ready for independently executing the control action (\ref{eq: ui}).

\section{Formation motion control}
\label{sec: comp}
\subsection{Translation and rotation of the robotic team}
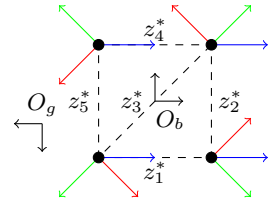
\begin{figure}
\centering
\begin{tikzpicture}[line join=round]
\draw[dashed](0,0)--(1.5,0)--(1.5,1.5)--(0,1.5)--(0,0)--(1.5,1.5);
\draw[draw=red,arrows=->](0,0)--(.525,-.525);
\draw[draw=red,arrows=->](1.5,0)--(2.025,.525);
\draw[draw=red,arrows=->](1.5,1.5)--(.975,2.025);
\draw[draw=red,arrows=->](0,1.5)--(-.525,.975);
\draw[draw=blue,arrows=->](0,0)--(.75,0);
\draw[draw=blue,arrows=->](1.5,0)--(2.25,0);
\draw[draw=blue,arrows=->](1.5,1.5)--(2.25,1.5);
\draw[draw=blue,arrows=->](0,1.5)--(.75,1.5);
\draw[draw=green,arrows=->](0,0)--(-.525,-.525);
\draw[draw=green,arrows=->](1.5,0)--(2.025,-.525);
\draw[draw=green,arrows=->](1.5,1.5)--(2.025,2.025);
\draw[draw=green,arrows=->](0,1.5)--(-.525,2.025);
\filldraw(0,0) circle (2pt);
\filldraw(1.5,0) circle (2pt);
\filldraw(1.5,1.5) circle (2pt);
\filldraw(0,1.5) circle (2pt);
\draw[draw=black,arrows=->](.75,.75)--(1.125,.75);
\draw[draw=black,arrows=->](.75,.75)--(.75,1.125);
\draw[draw=black,arrows=->](-.75,.45)--(-1.125,.45);
\draw[draw=black,arrows=->](-.75,.45)--(-.75,.075);
\node at (.95,.5) {\small $O_b$ \normalsize};\node at (-.75,.7) {\small $O_g$ \normalsize};\node at (.75,-.2) {\small $z_1^*$ \normalsize};\node at (1.75,.75) {\small $z_2^*$ \normalsize};\node at (.45,.75) {\small $z_3^*$ \normalsize};\node at (.75,1.7) {\small $z_4^*$ \normalsize};\node at (-.25,.75) {\small $z_5^*$ \normalsize};\end{tikzpicture}
	\caption{According to (\ref{eq: uid}), the velocity of the robots at the desired shape is the linear combination of the unit vectors from their associated relative positions (black dashed lines). The velocity of the robots can be split in the superposition of several ones as shown in (\ref{eq: sigma}). For example, the translational velocity for the shape is marked in blue color and it can be constructed by multiplying $z_1^*$ or $z_4^*$ by the motion parameters $\sigma^{(t1)}_1$ and $\sigma^{(t1)}_4$ respectively. In fact, the bigger the motion parameter, the higher the speed of the translational motion. It is straightforward to see that following this procedure, one can construct rotational motion (red velocity vectors), and even scaling motion (green velocity vectors). Obviously, these velocity vectors are always with respect to a frame of coordinates $O_b$ fixed with the desired shape, and not with a fixed frame of global coordinates $O_g$. Making this approach independent of any external localization system.}
\label{fig: mis}
\end{figure}

The control action (\ref{eq: ui}) only leads the team of robots to achieve the static formation of a rigid shape. We further need to extend such a control action in order to induce a collective motion, e.g., translations and rotations. A novel algorithm that assign extra velocities to the robots based on their relative positions with respect to their neighbors, allows to achieve such a desired collective motions \cite{de2016distributed}. Even with this technique one can control the scaling of the shape \cite{de2016distributedb} while guaranteeing stability and convergence properties. In particular, let us focus only in the control of the inter-robot distance for the edge $\mathcal{E}_k=(i,j)$, the extension of (\ref{eq: ui}) only for such an edge can be written as follows in two terms
\begin{equation}
	^iu_i^{(k)} = -\frac{^i(p_i-p_j)}{||p_i-p_j||}(||p_i-p_j|| - d_{(i,j)}) + \sigma_k \frac{^i(p_i-p_j)}{||p_i-p_j||}, \label{eq: uim}
\end{equation}
where $\sigma_k \in\mathbb{R}$ is a motion parameter associated to its corresponding edge. While the first term in (\ref{eq: uim}) controls the inter-robot distance, the second one adds an extra velocity which clearly depends on the current relative position between neighboring robots. The parameter $\sigma_k$ can be chosen small enough\footnote{Conversely, a gain for the controller for the error distances can be set big enough.} such that the first term achieves the objective of driving the error distance $e_k$ to zero. In such a situation we achieve the following equality
\begin{equation}
	^iu_i^{*(k)} = \sigma_k \frac{^i(p_i-p_j)^*}{||(p_i-p_j)^*||}, \label{eq: uid}
\end{equation}
where the superscript $*$ indicates that the relative positions of the robots describe the desired shape, i.e., once all the $e_k$ are equal to zero. Note that in such a case, if we take into account all the edges associated to a robot, the vector velocity of the robot is a linear combination of its relative positions as it is illustrated in Figure \ref{fig: mis}. If the parameters $\sigma_k$ are chosen appropriately, they can describe rotations and translations of the desired shape \cite{de2016distributed}. Therefore, the second term in (\ref{eq: uim}) does not brake the desired shape. In fact, $\sigma_k$ can be split in
\begin{equation}
	\sigma_k = \sigma_k^{(t_1)} + \sigma_k^{(t_2)} + \sigma_k^{(r)}, \label{eq: sigma}
\end{equation}
where the superscripts denote for different translations (vertical and horizontal) and rotations. For example, we can assign to one of the axis of a joystick the possibility to activate or inhibit one of the parameters in (\ref{eq: sigma}), so an operator can easily control the whole formation movement.

\subsection{Experimental validation}
\begin{figure}
\centering
\begin{subfigure}{.2\textwidth}
  \centering
  \includegraphics[width=\linewidth]{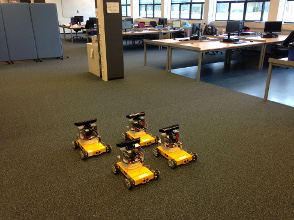}
  \caption{Initial position ($t=0$).}
  \label{fig:RC_1}
\end{subfigure}%
\hspace{.5pt}
\begin{subfigure}{.2\textwidth}
  \centering
  \includegraphics[width=\linewidth]{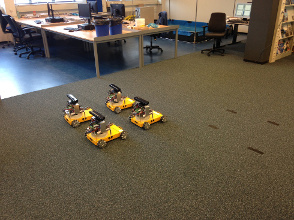}
  \caption{Moving diagonally with the desired shape ($t=20$).}
  \label{fig:RC_3}
\end{subfigure} 
\\
\begin{subfigure}{.2\textwidth}
  \centering
  \includegraphics[width=\linewidth]{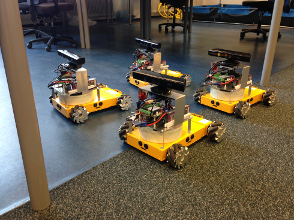}
  \caption{Between furniture ($t=35$).}
  \label{fig:RC_4}
\end{subfigure}%
\hspace{.5pt}
\begin{subfigure}{.2\textwidth}
  \centering
  \includegraphics[width=\linewidth]{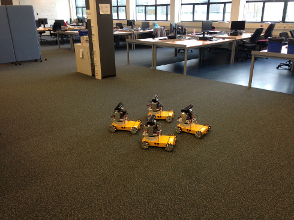}
  \caption{Back at initial positions in a rotated configuration of the desired shape ($t=147$).}
  \label{fig:RC_9}
\end{subfigure}
\caption{Pictures of the formation travelling around in an office space.}
\label{fig:RC}
\end{figure}

\begin{figure*}
\centering
\begin{subfigure}[t]{.25\textwidth}
  \centering
  \includegraphics[width=\linewidth]{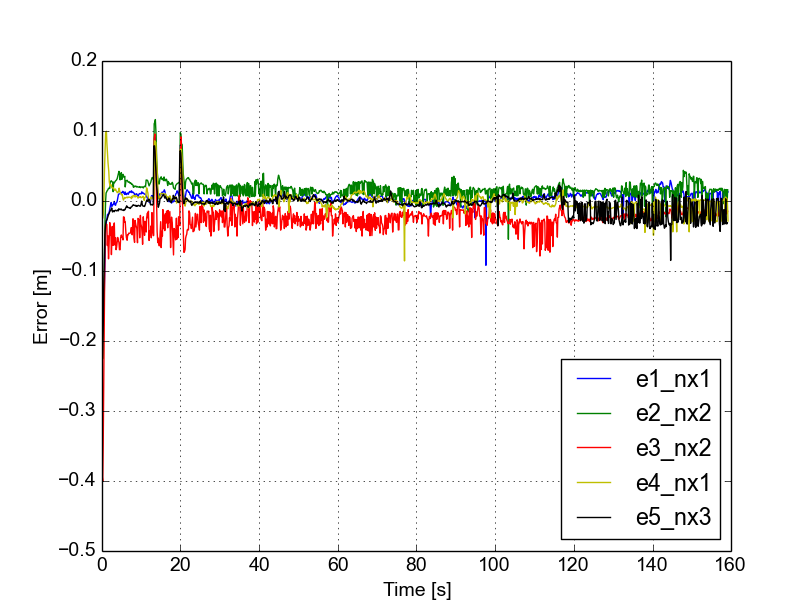}
	\caption{Inter-robot distance errors measured from the positive side of the edge as defined in (\ref{eq: B}).}
  \label{fig:exp71}
\end{subfigure}%
\begin{subfigure}[t]{.25\textwidth}
  \centering
  \includegraphics[width=\linewidth]{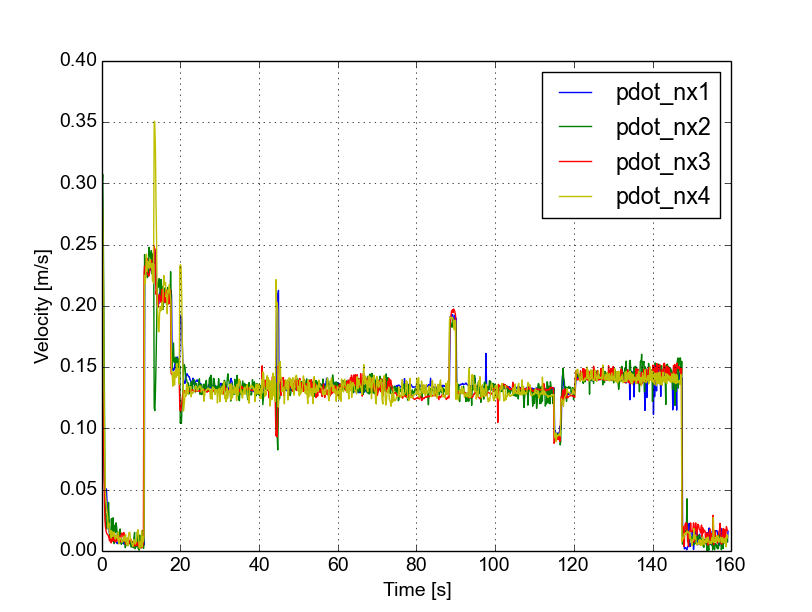}
  \caption{Speed of the robots.}
  \label{fig:exp73}
\end{subfigure}
\begin{subfigure}[t]{.25\textwidth}
  \centering
  \includegraphics[width=\linewidth]{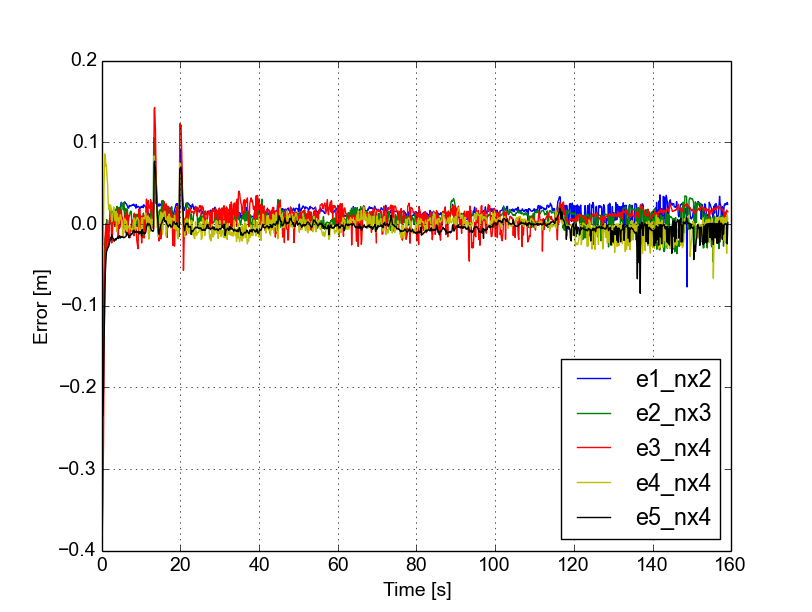}
	\caption{Inter-robot distance errors measured from the negative side of the edge as defined in (\ref{eq: B}).}
  \label{fig:exp72}
\end{subfigure}
	\caption{Note how the error signals differ depending on which neighboring robot takes the measurement.}
\label{fig:exp7}
\end{figure*}

We present an experiment with a team of four robots as described in Section \ref{sec: platform} travelling around in a squared formation between furniture in an office by executing the control law (\ref{eq: uid}). The video associated to this experiment can be watched at https://www.youtube.com/watch?v=qdkDreHntNk, and captions are displayed in Figure \ref{fig:RC}. An operator gives a sequence of commands with a joystick for translating and rotating the formation whenever he desires to change the course of the team. The rest of the time, there is total radio silence since the robots only rely on their laser scanners. The plots in Figure \ref{fig:exp7} show how the formation translates around with the given constant velocities with \emph{practically} a perfect shape as the error signals indicate. However, we can notice that the errors do not convergence asymptotically to zero. In fact, depending on which neighboring robot we look at for a particular edge, the error signal for the same distance to be controlled differs. This is the anticipated issue due to the non-calibrated laser scanners. Consequently, the formation does not follow precisely the commanded translation given by the operator, who has to regularly correct the course of the formation. Even when the laser scanners only differ in few millimeters for a squared shape of side of a meter. We will see in the next subsection \ref{sec: cal} that this issue can be eliminated with an online calibration, that can be executed at the same time while the formation is moving.

\subsection{Online calibration}
\label{sec: cal}

In our work in \cite{garcia2015controlling} we propose an online estimator in order to compensate distance disagreements between neighboring robots. In particular, it considers the following situation for two neighboring agents $i$ and $j$ sharing the same edge $\mathcal{E}_k$
\begin{equation}
\begin{cases}
u^{(k)}_i =  -\hat z_k \big(e_k - \mu_k)\\
u^{(k)}_j =   \hat z_k e_k.
\end{cases}
\label{eq: ukib}
\end{equation}
Note that the selection of the error signal (\ref{eq: e}) makes trivial to identify $\mu_k$ as a bias factor in the range reading $||z_k||$ (instead of a disagreement on $d_k$). We highlight that this is not the case when the error signal is the usual one considered in the literature, i.e., $e_k = ||z_k||^2 + d_k^2$ like in our work in \cite{garcia2015controlling}, since the squared signals.

\begin{figure}
\centering
\begin{subfigure}{.49\columnwidth}
  \centering
  \includegraphics[width=\linewidth]{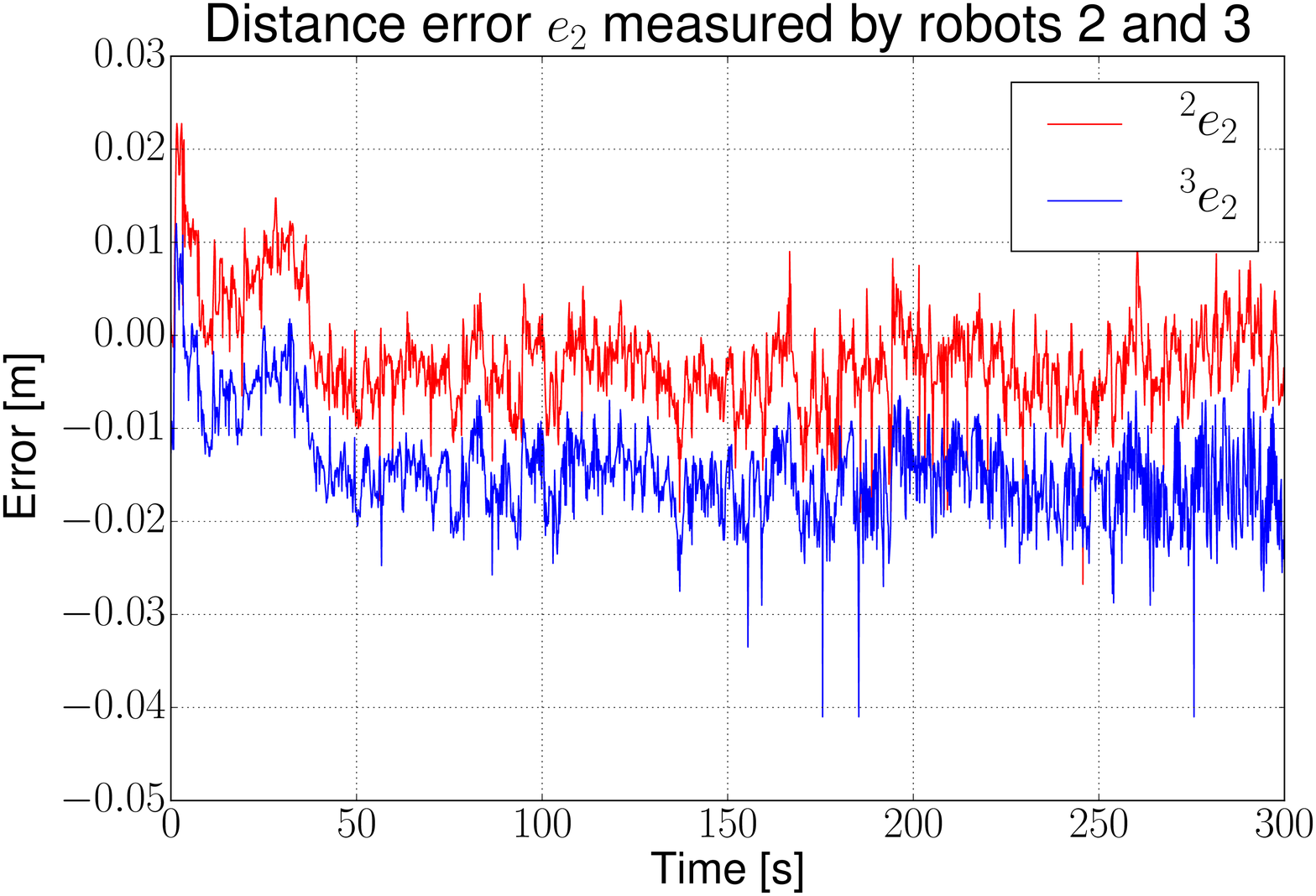}
\end{subfigure}
\begin{subfigure}{.49\columnwidth}
  \centering
  \includegraphics[width=\linewidth]{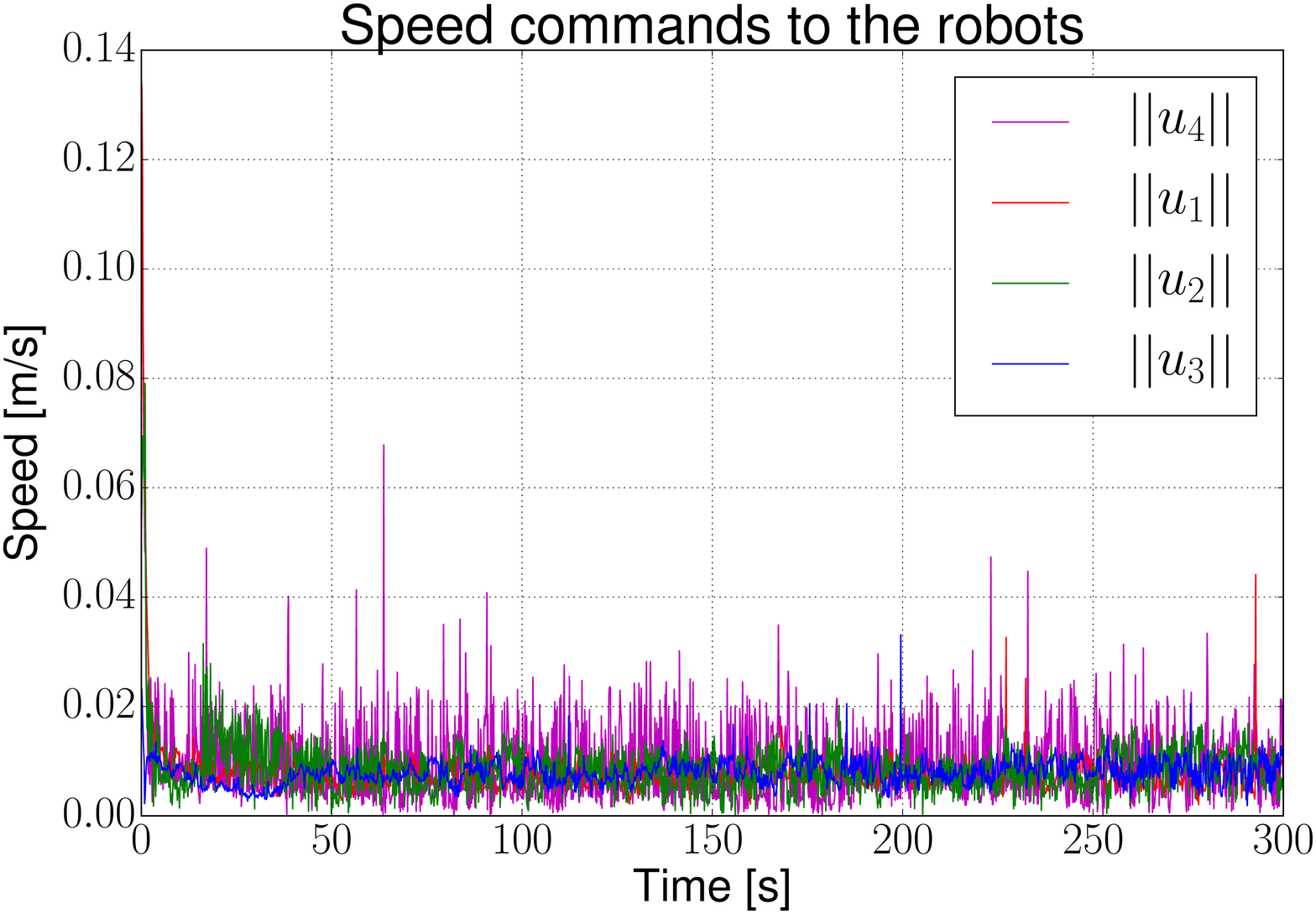}
\end{subfigure}
	\caption{Small biases (of the order of millimeters) in the range sensors between robots lead to different error signals $e_k$ at each robot. As a result, the control actions converge to a non-zero mean value. Consequently, an undesired motion of the formation occurs at the steady-state. The large spikes in the signals are due to the fact that the laser scanners instead of hitting the other sensors sometimes hit another parts of the neighboring robots.}
\label{fig: mu1}
\end{figure}

\begin{figure}
\centering
\begin{subfigure}{.49\columnwidth}
  \centering
  \includegraphics[width=\linewidth]{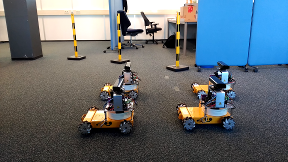}
  \caption{0 secs.}
\end{subfigure}
\begin{subfigure}{.49\columnwidth}
  \centering
  \includegraphics[width=\linewidth]{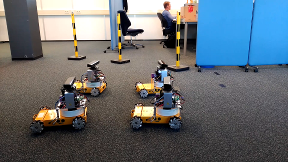}
  \caption{300 secs.}
\end{subfigure}
	\caption{The formation achieves quickly in three seconds the desired one meter side square. However, due to discrepancies of the order of millimeters between the range sensors, there is a undesired collective motion that it is noticeable after a sufficiently long time.}
\label{fig: mevoy}
\end{figure}

For the edge $\mathcal{E}_k = (i,j)$ we have two different range sensors, one at each neighboring robot, measuring $||z_k||$. Only a discrepancy, or a biased sensor, in one of the edges is needed for \emph{pulling} the rest of the formation \cite{mou2016undirected}. This has consequences as we have explained in the previous experiment, where the shape is not perfectly achieved, and the operator has to correct the course of the team very often due to an undesired superposed motion. The employed laser scanners have roughly a precision of millimeters without being calibrated out of the box. \emph{A priori} it sounds good enough for target distances of about a meter or more. We have that the corresponding control actions for the robots in the edge $\mathcal{E}_k=(i,j)$ are
\begin{equation}
\begin{cases}
	u^{(k)}_i &=  -\hat z_k \big((||z_k|| - \mu_k) - d_k\big) \\
	u^{(k)}_j &=  \hat z_k (||z_k|| - d_k),
\end{cases} 
\label{eq: uki}
\end{equation}
where $\mu_k\in\R$ is now the constant bias between the laser scanners of robots $i$ and $j$. Note that the measured range by robot $i$ is given by $(||z_k|| + \mu_k)$ as a whole, and since there is not communication with its neighbor, it cannot trivially figure out the value of $\mu_k$. We run an experiment with a static shape in order to measure the impact of this bias. At the beginning of the experiment the robots are collocated close to the desired shape, and after a short transition time the error signals do not converge to zero since the presence of the biases $\mu_k$. This is illustrated in Figure \ref{fig: mu1}, where as an example the magnitude of $\mu_1$ is estimated to be around six millimeters. Consequently, the control action of the robots converge to a \emph{non-zero mean} stationary value. The robots cannot track a speed below around $1.5$ cm/s because of the friction with the floor. Nevertheless, because the steady-state control signal is closer to that threshold than to zero, due to the noise in the laser scanner, spikes bigger than the threshold are more likely to occur. Since there is a privileged direction for the undesired motion of the formation \cite{mou2016undirected,de2016distributed}, this one will change its location after a sufficient long time as it is illustrated in Figure \ref{fig: mevoy}. In this case, about a meter after five minutes.

\begin{figure}
\centering
\begin{subfigure}{.49\columnwidth}
  \centering
  \includegraphics[width=\linewidth]{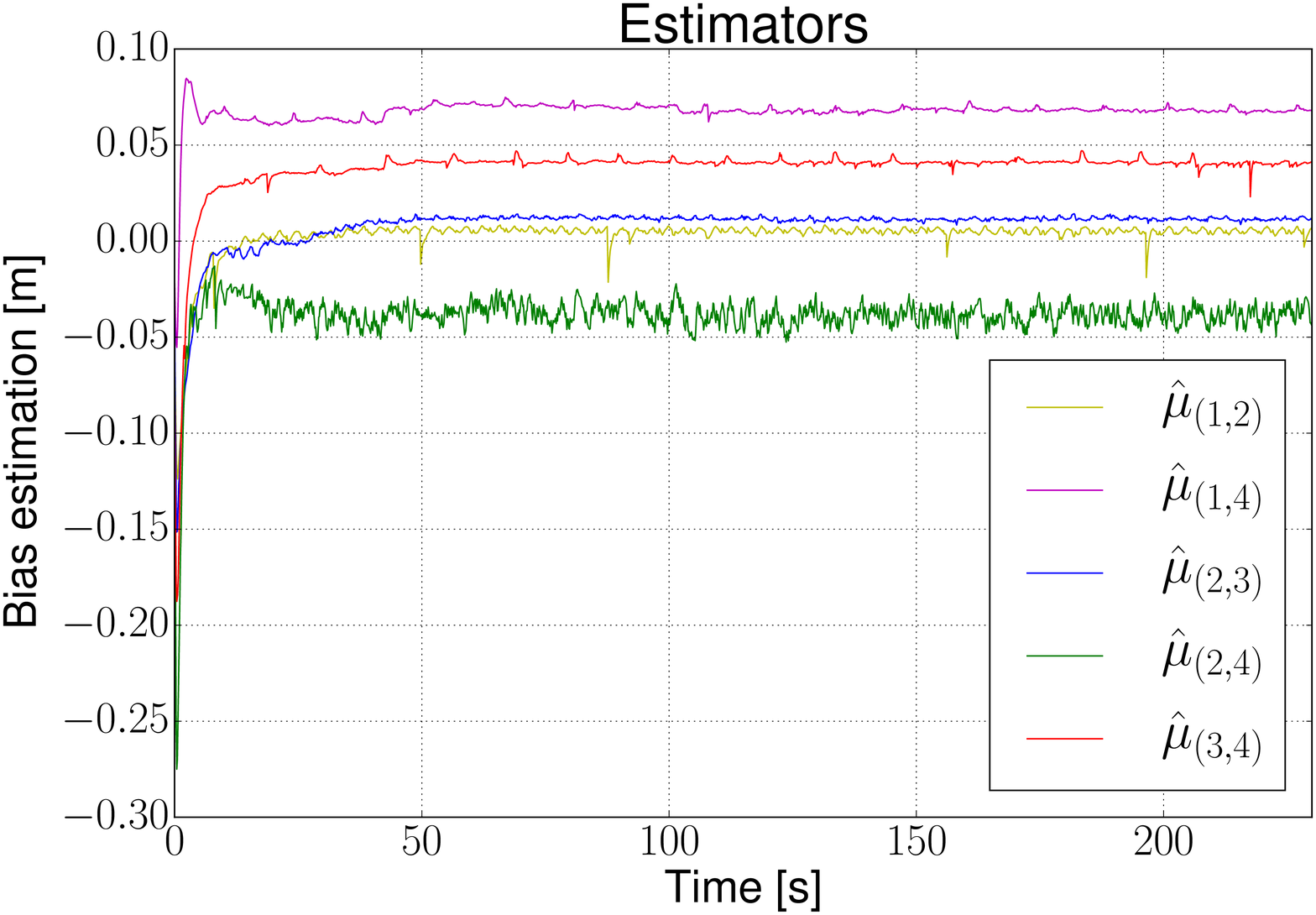}
\end{subfigure}
\begin{subfigure}{.49\columnwidth}
  \centering
  \includegraphics[width=\linewidth]{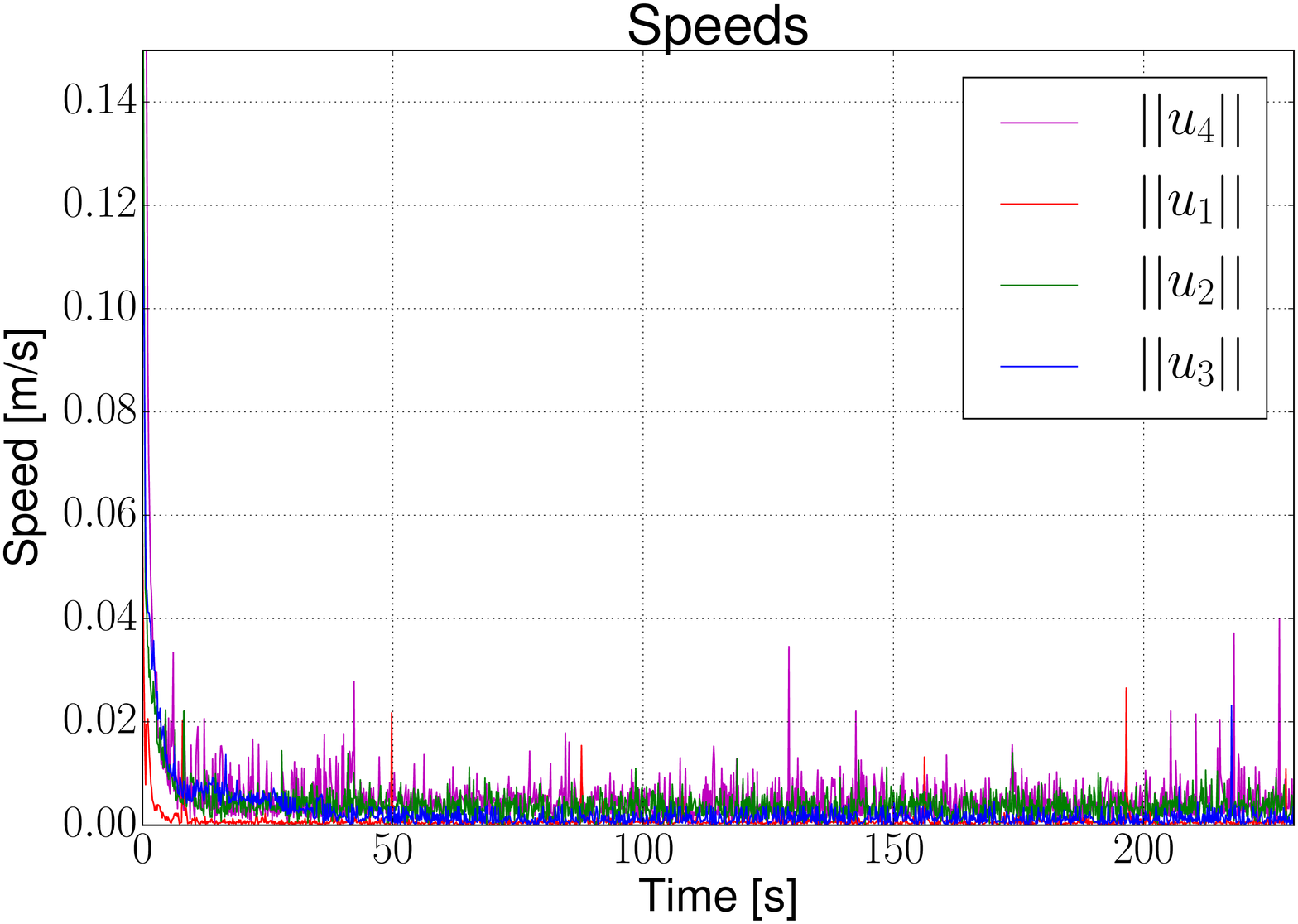}
\end{subfigure}
	\caption{The robots estimate their biases $\mu_k$ which helps to the convergence of the control error signal to zero where the speeds are also practically, i.e., the robots eventually are at the desired distances with respect to its neighbors. In this experiment, the gain $c$ of the estimator (\ref{eq: muhat}) has been set to $1$.}
\label{fig: estimators}
\end{figure}

We choose only one robot to estimate the discrepancy $\mu_k$ with respect to its neighbor. The dynamics of such an estimator \cite{garcia2015controlling} and its integration for the robot $i$ estimating the discrepancy $\mu_{(i,j)}$ in the edge $\mathcal{E}_k = (i,j)$ is given by
\begin{equation}
	\begin{cases}
		^iu_i^{(k)} = - \frac{^i(p_i-p_j)}{||p_i-p_j||}(^i||p_i-p_j|| - d_{(i,j)} - \hat\mu_{(i,j)}) \\
		\dot{\hat\mu}_{(i,j)} = c(^i||p_i-p_j|| - d_{(i,j)} - \hat\mu_{(i,j)}),
	\end{cases}
	\label{eq: muhat}
\end{equation}
where $c\in\R^+$ is a sufficiently big gain in order to make the estimator to converge quickly, so the gradient descent control for the formation behaves as expected. The selection of the estimating robots is not arbitrary \cite{garcia2015controlling}. In fact, this task can be represented by a directed graph, where the tails of the arrows indicate which robot is estimating the discrepancy of a given edge. It suffices that such a graph does not contain any loops. The algorithm (\ref{eq: muhat}) guarantees the local exponential convergence of the estimators to the unknown biases, which in practice means to have an online calibration of the sensors. As a result, the formation achieves the commanded desired shape and will not exhibit any undesired motion as it can be seen in Figure \ref{fig: estimators}. Indeed, the online calibration (\ref{eq: muhat}) can be used at the same time than the motion control (\ref{eq: uid}). Therefore, improving the practical operation of the motion of the formation by requiring less corrections from the external operator.

\section{Conclusions}
\label{sec: con}
The goal of the presented research has been to show the design and implementation of a robust fully distributed motion formation control algorithm for a team of mobile robots. We have experimentally validated that the formation control algorithms based on undirected graphs cannot be correctly executed if the sensors in neighboring robots are not perfectly calibrated. Although this issue was predicted theoretically in \cite{mou2016undirected}, the real practical impact in a fully distributed system had not been investigated yet. We have shown that the algorithm presented in \cite{garcia2015controlling} can also be employed for online sensor calibration by appropriately defining the error distance signal. We have further shown that for a common setup of mobile robots with laser scanners, no other issues in practice arise provoking unexpected behaviors. In fact, the proposed motion algorithm and its implementation can be run in an inexpensive microcontroller, and it is robust for driving the formation around obstacles like office furniture.

\bibliographystyle{IEEEtran}
\bibliography{hector_ref.bib}

\end{document}